\documentclass[letterpaper]{article}
\usepackage{proceed2e}
\usepackage[utf8]{inputenc}
\usepackage[margin=1in]{geometry}

\usepackage{times}

\usepackage{amsmath}
\usepackage{amssymb,mathtools}
\usepackage{natbib}
\usepackage{hyperref}
\usepackage{xcolor}

\usepackage{placeins}

\usepackage{tikz, pgfplots}
\pgfplotsset{compat=newest}
\usetikzlibrary{plotmarks, pgfplots.colormaps, pgfplots.groupplots}
\newlength\figurewidth
\newlength\figureheight
\pgfkeys{/pgfplots/mystyle/.style={
  legend style = {fill=none, draw=none}, %, at={(0.01, 0.01)}},
  tick style={major tick length=4pt,semithick,gray},
  xtick align = inside,
  ytick align = inside,
  group style = {vertical sep = 0cm},
  legend columns = 1
  }}
\definecolor{lred}{RGB}{200,0,0}
\definecolor{dred}{RGB}{130,0,0}
\definecolor{ldre}{RGB}{190.0515  127.5000  127.5000}

\definecolor{dblu}{RGB}{0,0,130}
\definecolor{lblu}{RGB}{127.5000, 127.5000, 192.3975}

\definecolor{dgre}{RGB}{0,130,0} 
\definecolor{dgra}{RGB}{50,50,50}
\definecolor{mgra}{RGB}{100,100,100}
\definecolor{lgra}{RGB}{220,220,220}

\definecolor{MPG}{RGB}{000,125,122}

\usepackage{algorithm}
\usepackage{algpseudocode}
\makeatletter
\def\therule{\makebox[\algorithmicindent][l]{\hspace*{.5em}\vrule height .75\baselineskip depth .25\baselineskip}}%

\newtoks\therules% Contains rules
\therules={}% Start with empty token list
\def\appendto#1#2{\expandafter#1\expandafter{\the#1#2}}% Append to token list
\def\gobblefirst#1{% Remove (first) from token list
  #1\expandafter\expandafter\expandafter{\expandafter\@gobble\the#1}}%
\def\LState{\State\unskip\the\therules}% New line-state
\def\pushindent{\appendto\therules\therule}%
\def\popindent{\gobblefirst\therules}%
\def\printindent{\unskip\the\therules}%
\def\printandpush{\printindent\pushindent}%
\def\popandprint{\popindent\printindent}%

\algdef{SE}[WHILE]{While}{EndWhile}[1]
  {\printandpush\algorithmicwhile\ #1\ \algorithmicdo}
  {\popandprint\algorithmicend\ \algorithmicwhile}%
\algdef{SE}[FOR]{For}{EndFor}[1]
  {\printandpush\algorithmicfor\ #1\ \algorithmicdo}
  {\popandprint\algorithmicend\ \algorithmicfor}%
\algdef{S}[FOR]{ForAll}[1]
  {\printindent\algorithmicforall\ #1\ \algorithmicdo}%
\algdef{SE}[LOOP]{Loop}{EndLoop}
  {\printandpush\algorithmicloop}
  {\popandprint\algorithmicend\ \algorithmicloop}%
\algdef{SE}[REPEAT]{Repeat}{Until}
  {\printandpush\algorithmicrepeat}[1]
  {\popandprint\algorithmicuntil\ #1}%
\algdef{SE}[IF]{If}{EndIf}[1]
  {\printandpush\algorithmicif\ #1\ \algorithmicthen}
  {\popandprint\algorithmicend\ \algorithmicif}%
\algdef{C}[IF]{IF}{ElsIf}[1]
  {\popandprint\pushindent\algorithmicelse\ \algorithmicif\ #1\ \algorithmicthen}%
\algdef{Ce}[ELSE]{IF}{Else}{EndIf}
  {\popandprint\pushindent\algorithmicelse}%
\algdef{SE}[PROCEDURE]{Procedure}{EndProcedure}[2]
   {\printandpush\algorithmicprocedure\ \textproc{#1}\ifthenelse{\equal{#2}{}}{}{(#2)}}%
   {\popandprint\algorithmicend\ \algorithmicprocedure}%
\algdef{SE}[FUNCTION]{Function}{EndFunction}[2]
   {\printandpush\algorithmicfunction\ \textproc{#1}\ifthenelse{\equal{#2}{}}{}{(#2)}}%
   {\popandprint\algorithmicend\ \algorithmicfunction}%
\makeatother

\title{Coupling Adaptive Batch Sizes with Learning Rates}

\author{ {\bf Lukas Balles} \\
MPI for Intelligent Systems \\
T\"ubingen, Germany \\
\And
{\bf Javier Romero}\thanks{\hspace{6pt}Work done while at MPI for Intelligent Systems.}  \\
Body Labs Inc.          \\
New York, NY, USA \\
\And
{\bf Philipp Hennig}   \\
MPI for Intelligent Systems \\
T\"ubingen, Germany \\
}

\begin{document}

\maketitle

\begin{abstract}
Mini-batch stochastic gradient descent and variants thereof have become standard for large-scale empirical risk minimization like the training of neural networks.
These methods are usually used with a constant batch size chosen by simple empirical inspection. 
The batch size significantly influences the behavior of the stochastic optimization algorithm, though, since it determines the variance of the gradient estimates.
This variance also changes over the optimization process; when using a constant batch size, stability and convergence is thus often enforced by means of a (manually tuned) decreasing learning rate schedule.

We propose a practical method for dynamic batch size adaptation. It estimates the variance of the stochastic gradients and adapts the batch size to decrease the variance proportionally to the value of the objective function, removing the need for the aforementioned learning rate decrease.
In contrast to recent related work, our algorithm couples the batch size to the learning rate, directly reflecting the known relationship between the two.
On popular image classification benchmarks, our batch size adaptation yields faster optimization convergence, while simultaneously simplifying learning rate tuning.
A TensorFlow implementation is available.
\end{abstract}

\section{INTRODUCTION}

In parametric machine learning models, like logistic regression or neural networks, the performance of a parameter vector $w\in\mathbb{R}^d$ on datum $x$ is quantified by a loss function $\ell(w;x)$.
Assuming the data comes from a distribution $x\sim p$, the goal is to minimize the expected loss, or \emph{risk},
\begin{equation}
\mathcal{R}(w) = \mathbf{E}_{x\sim p} [ \ell(w; x) ].
\end{equation} 
We consider \emph{empirical risk minimization} tasks of the form
\begin{equation}
\label{eq:erm_problem}
\min_{w\in\mathbb{R}^d} F(w) = \frac{1}{M} \sum_{i=1}^M \ell(w; x_i),
\end{equation}
where the risk is approximated using a training set $\{x_1, \dotsc, x_M\}$ of data sampled (approximately) from $p$.
Typical optimization algorithms used to minimize \eqref{eq:erm_problem} repeatedly evaluate the gradient
\begin{equation}
\label{eq:exact_gradient}
\nabla F(w) = \frac{1}{M} \sum_{i=1}^M \nabla \ell(w; x_i).
\end{equation}
For large-scale problems where $M$ and/or $d$ are large, it is inefficient or impossible to evaluate the exact gradient \eqref{eq:exact_gradient}, and one typically resorts to stochastic gradients by randomly drawing a mini-batch $\mathcal{B} \subset \{1,\dotsc, M\}$, $\vert \mathcal{B}\vert =m \ll M$, at each step of the optimization algorithm and using the gradient approximation
\begin{equation}
g(w) = \frac{1}{m} \sum_{i\in\mathcal{B}} \nabla \ell(w; x_i).
\end{equation}
The simplest, but still widely used, stochastic optimization algorithm is stochastic gradient descent \citep[\textsc{sgd}, ][]{Robbins1951}, which updates
\begin{equation}
w_{k+1} = w_k - \alpha_k g(w_k),
\end{equation}
where $\alpha_k \in\mathbb{R}_+$ is the step size parameter, often called \emph{learning rate} in the machine learning context.
Variants of \textsc{sgd} include \textsc{adagrad} \citep{Duchi2011}, \textsc{adadelta} \citep{Zeiler2012}, and \textsc{adam} \citep{Kingma2014}.
We restrict our considerations to \textsc{sgd} in this paper.
%, but the method is transferable to other algorithms with minor adaptations.

\subsection{THE EFFECT OF THE BATCH SIZE}
\label{variance}

If $i$ is drawn uniformly at random from $\{ 1,\dotsc, M\}$, $\nabla \ell_i(w) = \nabla \ell(w; x_i)$ is a random variable with mean
\begin{equation}
\mathbf{E}[\nabla \ell_i(w) ] = \frac{1}{M} \sum_{i=1}^M \nabla \ell_i(w) = \nabla F(w), 
\end{equation}
and covariance matrix
\begin{equation}
\label{eq:exact_variance}
\begin{split}
& \Sigma(w) := \mathbf{cov} \left[ \nabla \ell_i(w) \right] = \\
& \frac{1}{M}\sum_{i=1}^M (\nabla \ell_i(w) - \nabla F(w))(\nabla \ell_i(w) - \nabla F(w))^T.
\end{split}
\end{equation}
Likewise, a stochastic gradient $g(w)$ computed on a randomly-drawn mini-batch $\mathcal{B}$ is a random variable with mean $\nabla F(w)$.
Assuming that it is composed of $m$ samples drawn independently with replacement, its covariance matrix is
\begin{equation}
\mathbf{cov} [ g(w) ] = \frac{\Sigma(w)}{m}
\end{equation}
and, by the Central Limit Theorem, $g(w)$ is approximately normally distributed:
\begin{equation}
\label{eq:gaussian_assumption}
g(w) \sim \mathcal{N}\left( \nabla F(w), \frac{\Sigma(w)}{m} \right).
\end{equation}
When sampling without replacement, as is usually done in practice, the same holds \emph{approximately} as long as $m\ll M$.

In practice, the batch size $m$ is often set to a fixed value, which is chosen ad hoc or by simple empirical tests. But it is actually a crucial variable, which poses an intricate trade-off that affects the optimizer's performance. On the one hand, the variance of the stochastic gradients \emph{decreases linearly} with $m$, so small batches give vague gradient information, thus slow convergence in the number of optimization steps. On the other hand, the cost per step \emph{increases linearly} with $m$. (This assumes that batch sizes are large enough to fully utilize the available parallel computing resources, which can easily be guaranteed by enforcing an appropriate minimal batch size.) While we can thus linearly trade off variance and cost, the gradient variance does \emph{not} linearly affect the performance of the optimizer; its effect depends on the local structure of the objective and interacts with other parameter choices of the optimizer, notably the learning rate. In general, there should thus be an \emph{optimal} batch size that balances these two aspects. Choosing such good batch sizes is an important aspect in the design of a numerical optimizer.

Below, we propose an algorithm that adapts the batch size based on the gradient variance observed by the optimizer at runtime. 
The exact variance  over the entire data set \eqref{eq:exact_variance} is prohibitively costly to compute, but it can be estimated by the sample variance computed on a mini-batch.
As will be described below, we only require the diagonal elements of $\Sigma(w)$, corresponding to the variances of the individual components.
These can be estimated by
\begin{equation}
\label{eq:sample_variance}
S(w) = \frac{1}{m} \sum_{i\in\mathcal{B}} \nabla \ell_i(w)^{.2} - g(w)^{.2} \in \mathbb{R}^d,
\end{equation}
where $^{.2}$ signifies an element-wise square.

\subsection{NOTATION}

The following discussion addresses the choice of batch size for a single \textsc{sgd} step, assuming that we are currently at some arbitrary but fixed point $w$ in parameter space.
For notational convenience, we will thus drop $w$ from the notation and write $F=F(w)$, $\nabla F = \nabla F(w)$, et cetera.

\section{RELATED WORK}

The dynamic adaption of batch sizes has already attracted attention in other recent works.
\cite{Friedlander2012} derive decreasing series of bounds on the gradient variance that provably yield fast convergence rates with a \emph{constant} learning rate, showing that an increasing batch size can replace a decreasing learning rate. 
To realize these bounds in practice, they propose to increase the batch size by a pre-specified constant factor in each iteration, without adaptation to (an estimate of) the gradient variance.

The prior works closest to ours in spirit are by \cite{Byrd2012} and \cite{De2016}, who propose to adapt the batch size based on variance estimates.
Their criterion is based on the observation that $-g$ is a descent direction if
\begin{equation}
\label{eq:descent_direction}
\Vert g - \nabla F\Vert \leq \theta \Vert g \Vert, \quad \text{ with } \quad 0 \leq \theta < 1
\end{equation}
(proof in Appendix \ref{appendix_mathematical_details}).
While the left-hand side of \eqref{eq:descent_direction} is of course unknown, one \emph{can} compute its expected square
\begin{equation}
\begin{split}
\mathbf{E}\left[ \Vert g - \nabla F\Vert^2 \right] & = \sum_{j=1}^d \mathbf{E}\left[ (g_j - \nabla F_j)^2 \right] \\
& = \sum_{j=1}^d \frac{\sigma_{jj}^2}{m} = \frac{\text{tr}(\Sigma)}{m}.
\end{split}
\end{equation}
Consequently, \eqref{eq:descent_direction} holds \emph{in expectation} if $\text{tr}(\Sigma)/m \leq \theta^2 \Vert g \Vert^2$ or (with equality)
\begin{equation}
\label{eq:noisy_grad_norm_rule}
m = \frac{1}{\theta^2} \frac{\text{tr}(\Sigma)}{\Vert g \Vert^2}.
\end{equation}
While this is a practical and intuitive method, the ``descent direction'' criterion is agnostic of the actual step being taken, which depends on the learning rate $\alpha$ in addition to the direction $-g$.
Moreover, the method introduces an additional free parameter $\theta$.
In this work we strive to alleviate these issues, while the resulting batch size adaptation rule will stay close to \eqref{eq:noisy_grad_norm_rule} in form and spirit.

A somewhat related line of research aims to reduce the variance of stochastic gradients by incorporating gradient information from previous iterations into the current gradient estimate.
Notable methods are \textsc{svrg} \citep{Johnson2013} and \textsc{saga} \citep{Defazio2014}.
Both are not mini-batch methods, since they update after gradient evaluations on individual training examples (which are then modified using stored gradient information). However, two recent papers \citep{Harikandeh2015, Daneshmand2016} combine these variance-reduced methods with increasing sample sizes, i.e., the effective size of the training set is increased over time.
In both, a sample size schedule has to be pre-specified and is not adapted at runtime.

We note that another recent line of work on non-uniform sampling of training samples with the goal of variance reduction \citep[including, but not limited to, ][]{Needell2014, Zhao2015, Schmidt2015, Csiba2016} is orthogonal to our work, since it is concerned with the composition of batches rather than their size.

More generally, our work fits into a recent effort to automate or simplify the tuning of parameters in stochastic optimization algorithms, most notably the learning rate \citep{Schaul2013, Mahsereci2015}.

\section{COUPLED ADAPTIVE BATCH SIZE}

We will cast the problem of finding a ``good'' batch size as maximizing a lower bound on the expected \emph{gain} per computational cost for an individual optimization step.
While the resulting rule is similar in form to \eqref{eq:noisy_grad_norm_rule}, it provides a new interpretation and introduces an explicit interaction with the learning rate.
This criterion will subsequently be simplified, removing all unknown quantities and free parameters from the equation.

\subsection{MAXIMIZING A BOUND ON THE EXPECTED GAIN}
\label{lipschitz_bound}

We define the \emph{gain} of the \textsc{sgd} step from $w$ to $w_+ = w-\alpha g$ as the drop in function value, $F - F_+$, where $F_+ = F(w_+)$.
In order to quantify this gain, we will assume that $F$ has Lipschitz-continuous gradients, i.e., there is a constant $L>0$ such that
\begin{equation}
\label{eq:lipschitz_property}
\Vert \nabla F(u) - \nabla F(v) \Vert \leq L \Vert u-v\Vert \quad \forall u, v\in\mathbb{R}^d.
\end{equation}
This is a standard assumption in the analysis of stochastic optimization algorithms, setting a not overly restrictive bound on how fast the gradient can change when moving in parameter space.
As a consequence, the change in $F$ from $v\in\mathbb{R}^d$ to $u\in\mathbb{R}^d$ is bounded (see, e.g., \cite{Bottou2016}, Eq.~4.3) by
\begin{equation}
\label{eq:lipschitz_bound}
F(u)  \leq F(v) + \nabla F(v)^T(u-v) + \frac{L}{2} \Vert	u-v\Vert^2.
\end{equation}
Inserting $v=w$ and $u = w_+ = w -\alpha g$ and rearranging yields a lower bound $\mathcal{G}$ on the gain:
%\begin{equation}
%F_+ \leq F - \alpha \nabla F^Tg + \frac{L\alpha^2}{2} \Vert g\Vert^2,
%\end{equation}
%and, after rearranging, implies a lower bound $\mathcal{G}$ on the gain:
\begin{equation}
\label{eq:gain_bound_definition}
F - F_+ \geq \mathcal{G} := \alpha \nabla F^T g - \frac{L\alpha^2}{2} \Vert g\Vert^2.
\end{equation}
To derive the expectation of $\mathcal{G}$, recall from Equation \eqref{eq:gaussian_assumption} that $\mathbf{E}[g] = \nabla F$ and
\begin{equation}
\label{eq:expectation_of_norm_g}
\begin{split}
\mathbf{E}\left[ \Vert g\Vert^2\right] &  = \sum_{j=1}^d \mathbf{E}\left[ g_j^2 \right]= \sum_{j=1}^d \left( \nabla F_j^2 + \frac{\sigma_{jj}^2}{m} \right) \\
& = \Vert \nabla F\Vert^2 + \frac{\text{tr}(\Sigma)}{m},
\end{split}
\end{equation}
where we used that, for $X\sim\mathcal{N}(\mu, \sigma^2)$, the second moment is $\mathbf{E}[X^2] = \mu^2 + \sigma^2$. 
Thus,
\begin{equation}
\label{eq:expected_gain}
\mathbf{E}[\mathcal{G}] = \left( \alpha - \frac{L \alpha^2}{2} \right) \Vert \nabla F\Vert^2  - \frac{L \alpha^2}{2m} \text{tr}(\Sigma).
\end{equation}
The first term in \eqref{eq:expected_gain} is the gain in absence of noise, determined by $\alpha$ and $\nabla F$.
It is reduced by a term that depends on the gradient variance and drops with $m$.
We see from \eqref{eq:expected_gain} that, for an expected descent, $\mathbf{E}[\mathcal{G}] > 0$, we require
\begin{equation}
\label{eq:alpha_max_with_noise}
\alpha < \frac{2\Vert \nabla F\Vert^2}{L\left( \Vert\nabla F\Vert^2 + \text{tr}(\Sigma)/m \right)},
\end{equation}
which exhibits a clear relationship between learning rate and batch size. Small batch sizes require a small learning rate, while larger batch sizes enable larger steps.
We will exploit this relationship later on by explicitly coupling the two parameters.
As a side note, for zero variance, we recover the well-known condition $\alpha < 2/L$ that guarantees convergence of gradient descent in the deterministic case.

Obviously, the larger $m$, the larger $\mathbf{E}[\mathcal{G}]$, so that the deterministic case is optimal if we ignore computational cost. Since that cost scales linearly with $m$, the optimal batch size is the one that maximizes expected gain per cost,
\begin{equation}
\label{eq:cost_sensitive_gain_maximization}
\max_m \, \frac{\mathbf{E}[\mathcal{G}]}{m}.
\end{equation}
A recent workshop paper \citep{Pirotta2016} used a similar idea, although on a different quantity (a statistical lower bound on the linearized improvement).
In our setting, maximal gain per cost is achieved by (derivation in Appendix \ref{appendix_mathematical_details})
\begin{equation}
\label{eq:lipschitz_bound_rule}
m = \frac{2L\alpha}{2-L\alpha} \frac{\text{tr}(\Sigma)}{\Vert \nabla F\Vert^2}.
\end{equation}

\subsection{THE CABS CRITERION}

The result in \eqref{eq:lipschitz_bound_rule} poses two practical problems. First, the Lipschitz constant $L$ is an unknown property of the objective function.
Even more importantly, it is difficult to reliably and robustly estimate the squared norm of the true gradient $\| \nabla F\|^2$ from a single batch.
One might be tempted to replace it with $\Vert g\Vert^2$, recovering a criterion similar to \eqref{eq:noisy_grad_norm_rule}, but this is not an unbiased estimator for the true gradient norm, as Equation \eqref{eq:expectation_of_norm_g} shows.
Depending on the noise level and, intriguingly, the batch size $m$, the second term in \eqref{eq:expectation_of_norm_g} can introduce a significant bias.

In an effort to address these practical problems, we propose to replace Eq.~\eqref{eq:lipschitz_bound_rule} with the following simpler rule, which we term the \emph{Coupled Adaptive Batch Size} (\textsc{cabs}):
\begin{equation}
\label{eq:CABS_rule}
m = \alpha \, \frac{\text{tr}(\Sigma)}{F}.
\end{equation}
A formal justification for this simplification will be given in \textsection\ref{mathematical_motivation_for_cabs}, but first we want to highlight some intuitive benefits of this batch size adaptation scheme.

A major advantage of the \textsc{cabs} rule, emphasized in its name, is the direct \emph{coupling} of learning rate and batch size.
We have established that a large learning rate demands large batches while a smaller, more cautious learning rate can be used with smaller batches (Equation \ref{eq:alpha_max_with_noise}).
The \textsc{cabs} rule explicitly reflects this known relationship.
Using \textsc{cabs} can thus be seen as ``tailoring'' the noise level to the learning rate the user has chosen.
We show experimentally, see \textsection\ref{experiments}, that this makes finding a well-performing learning rate easier.

Apart from that, theoretical considerations \citep{Friedlander2012} and experimental evidence show that it is beneficial to have small batches in the beginning and larger ones later in the optimization process.
Hence, one may want to think of the denominators of \eqref{eq:CABS_rule}, \eqref{eq:lipschitz_bound_rule} and \eqref{eq:noisy_grad_norm_rule} as a measure of ``optimization progress''.
The function value $F$ used in our \textsc{cabs} rule is, by definition, \emph{the} measure for training progress.
The norm of the true gradient $\Vert \nabla F \Vert^2$ conveys similar information (even though it might be misleading near non-optimal stationary points like saddle points or plateaus), but can not simply be estimated by $\Vert g\Vert^2$ as previously noted. We have also investigated unbiased estimators for $\Vert \nabla F\Vert^2$ by correcting the bias in $\Vert g\Vert^2$ using the variance estimate $S$, but these turned out to be too unreliable in experiments.
Additionally, Equation \eqref{eq:expectation_of_norm_g} also shows that using $\Vert g\Vert^2$ in the denominator leads to a disadvantageous feedback: larger batches cause $\Vert g\Vert^2$ to become smaller in expectation which, in turn, leads to larger batches according to \eqref{eq:noisy_grad_norm_rule} (and the other way round).

Readers who are rightly worried about the change of ``unit'' or ``type'' when replacing the gradient norm in \eqref{eq:lipschitz_bound_rule} with the function value in \eqref{eq:CABS_rule} may find it helpful to consider the units of measure for the quantities in \eqref{eq:CABS_rule}. Let $[w]$ and $[F]$ denote the units of the parameters and the objective function, respectively.
The gradient has unit $[F]/[w]$ and its variance $[F]^2/[w]^2$.
It is a key insight that a well-chosen learning rate has to be driven by quantities with unit $[w]^2/[F]$ (see \cite{MacKay2003} \textsection 34.4 for more discussion).
If this was not the case, the gradient descent update $- \alpha_k \nabla F(w_k)$ would not be \emph{covariant}, i.e., independent of the units of measure of $w$ and $F$.
It is also evident in Newton's method: in the one-dimensional case, a Newton update step is $- F^{\prime\prime}(w_k)^{-1} F^\prime(w_k)$, corresponding to a ``learning rate'' that is given by the inverse second derivative, having unit $[F]/[w]^2$.
Putting it all together, the right-hand side of \eqref{eq:CABS_rule} has unit
\begin{equation}
\frac{[w]^2}{[F]} \frac{[F]^2/[w]^2}{[F]} = [1].
\end{equation} 
Hence, the chosen batch size is invariant under rescaling of the objective.

Lastly, \textsc{cabs} realizes a bound on the gradient variance that is decreasing with the distance to optimality, similar to that in Theorem 2.5 of \cite{Friedlander2012}, which they have shown to guarantee convergence of \textsc{sgd} with a constant, non-decreasing learning rate.

\subsection{MATHEMATICAL MOTIVATION FOR CABS}
\label{mathematical_motivation_for_cabs}

For a more systematic motivation for the \textsc{cabs} rule, we will show that it is approximately equal to \eqref{eq:lipschitz_bound_rule}, and hence optimal in the sense of \eqref{eq:cost_sensitive_gain_maximization}, if we assume that $F$ locally has an approximately scalar Hessian, i.e.,
\begin{equation}
\label{eq:scalar_hessian}
\nabla^2 F(w) \approx hI, \quad h>0.
\end{equation}
First, note that under this assumption, the Lipschitz constant $L$ is exactly $h$ and the optimal batch size according to \eqref{eq:lipschitz_bound_rule} becomes
\begin{equation}
\label{eq:lipschitz_bound_rule_with_scalar_hessian}
m = \frac{2h\alpha}{2-h\alpha} \, \frac{\text{tr}(\Sigma)}{\Vert \nabla F \Vert^2}.
\end{equation}
Furthermore, the second-order Taylor expansion of $F$ around $w$ now reads
\begin{equation}
\label{eq:taylor_expansion_with_scalar_hessian}
F(u) \approx F(w) + \nabla F(w)^T (u-w) + \frac{h}{2} \Vert u-w\Vert^2.
\end{equation}
We minimize both sides with respect to $u$.
The left-hand side takes on the optimal value $F_\ast$.
For the right-hand side, we set the gradient with respect to $u$, $\nabla F(w) + h(u-w)$, to zero, which yields the minimizer $u=-\nabla F(w)/h + w$.
Inserting this back into \eqref{eq:taylor_expansion_with_scalar_hessian} and rearranging yields
\begin{equation}
\label{eq:grad_norm_to_distance_to_optimality}
\Vert \nabla F(w) \Vert^2 \approx 2h(F(w) - F_\ast).
\end{equation}
That is, we can replace the squared gradient norm with a scaled distance to optimality.
Doing so in \eqref{eq:lipschitz_bound_rule_with_scalar_hessian} reads
\begin{equation}
\label{eq:CABS_derivation_1}
m = \frac{\alpha}{2-h\alpha} \, \frac{\text{tr}(\Sigma)}{F - F_\ast}.
\end{equation}
We eliminate $h$ from this equation by realizing that, under the scalar Hessian assumption, a good learning rate is $\alpha=1/h$.
It corresponds both to the Newton step, as well as to the optimal constant learning rate $1/L$ for gradient descent, given that \eqref{eq:scalar_hessian} holds.
Hence, if we assume a well-chosen learning rate with $h\alpha \approx 1$, then Eq.~\eqref{eq:CABS_derivation_1} further simplifies to
\begin{equation}
\label{eq:CABS_with_bound}
m = \alpha \, \frac{\text{tr}(\Sigma)}{F-F_\ast}.
\end{equation}

Assuming a scalar Hessian is, of course, a substantial simplification. The result can partly be generalized to the less restrictive assumption of $\mu$-strong convexity, under which we still have (see Appendix \ref{appendix_mathematical_details})
\begin{equation}
\label{eq:grad_norm_to_distance_to_optimality_convexity}
\Vert \nabla F \Vert^2 \geq 2\mu (F - F_\ast),
\end{equation}
%Inserting this into \eqref{eq:lipschitz_bound_rule} yields
%\begin{equation}
%m \leq \frac{\alpha L}{(2-\alpha L)\mu} \frac{\text{tr}(\Sigma)}{F-F_\ast}.
%\end{equation}
If the problem is well-conditioned in that $\mu$ and $L$ are not too far from each other, the above argument carries through as an upper bound on the batch size.
%($2\mu > L$), we can choose appropriate step sizes ($\alpha \approx 2/L - 1/\mu$) that give rise to the CABS criterion.

To finally arrive at the \textsc{cabs} rule, we drop $F_\ast$.
This is based on the assumption of a non-negative loss, which holds for all standard loss functions like least-squares or cross-entropy.
In this case, $F\geq F-F_\ast$, i.e., the function value $F$ is a non-trivial upper bound on the distance to optimality.
If the optimum is close to zero, $F$ will be a good proxy for $F-F_\ast$.
If not, which is not uncommon, the denominator of the \textsc{cabs} rule has a small positive offset compared to \eqref{eq:CABS_with_bound}, but this will not fundamentally alter its implications, as long as we do not come too close to the optimum, which is usually the case for even modestly complex problems.
The more general form \eqref{eq:CABS_with_bound} can be used in lieu of \eqref{eq:CABS_rule} if one has access to a tighter lower bound on $F_\ast$, e.g., due to prior experience from similar problems.
In fact, when the objective function includes an additive regularization term, we suggest to use the unregularized loss as a proxy for $F-F_\ast$.

\section{PRACTICAL IMPLEMENTATION}

We outline a practical implementation of the \textsc{cabs} criterion.
Obviously, neither $F$ nor $\text{tr}(\Sigma)$ are known exactly at each individual \textsc{sgd} step, but estimates of both quantities can be obtained from a mini-batch.
This is straight-forward for the objective $F$. For the variance, we use the estimate $S$ explained in Equation \eqref{eq:sample_variance}.
Since $S$ only estimates the diagonal elements of the covariance matrix, it is $\text{tr}(\Sigma) \approx \Vert S\Vert_1$.
Considerations on how to practically compute $S$ can be found in \textsection\ref{variance_in_practice}.

\subsection{MECHANICAL DETAILS}
\label{CABS_in_practice}
We realize the \textsc{cabs} criterion in a predictive manner, meaning that we do \emph{not} find the exact batch size that satisfies \eqref{eq:CABS_rule} in each single optimization step.
To achieve such an exact enforcement of their criterion, \cite{Byrd2012} and \cite{De2016} increase the batch size by a small increment whenever the criterion is not satisfied, and only then perform the update.
This incremental computation introduces an overhead and, when the increment is small, can lead to under-utilization of computing resources.
Instead, we leverage the observation that gradient variance and function value change only slowly from one optimization step to the next, which allows us to use our current estimates of $F$ and $\text{tr}(\Sigma)$ to set the batch size used for the \emph{next} optimization step. It also allows for a smoothing of both quantities over multiple optimization steps. The estimates can be fairly noisy, especially that of $\text{tr}(\Sigma)$ at small batch sizes.
We use exponential moving averages (see Algorithm \ref{alg:CABS}) to obtain more robust estimates.

The resulting batch size is rounded to the nearest integer and clipped at minimal and maximal batch sizes.
A minimal batch size avoids under-utilization of the computational resources with very small batches and provides additional stability of the algorithm in the small-batch regime. 
A maximal batch size is necessary due to hardware limitations: In contemporary deep learning, GPU memory limits the number of samples that can be processed at once.
Our implementation has such a limit but it was never reached in our experiments.
We note in passing that \emph{algorithmic} batch size (the number of training samples used to compute a gradient estimate before updating the parameters) and \emph{computational} batch size (the number of training samples that are processed simultaneously) are in principle independent---a future implementation could split an algorithmic batch into feasible computational batches when necessary, freeing the algorithm from hardware-specific constraints. Algorithm \ref{alg:CABS} provides pseudo-code.
\begin{algorithm}
\caption{\textsc{sgd} with Coupled Adaptive Batch Size}
\label{alg:CABS}
\begin{algorithmic}[1]
\Require Learning rate $\alpha$, initial parameters $w_0$, number of steps $K$, batch size bounds $(m_\text{min}, m_\text{max})$, running average constant $\mu = 0.95$
\vspace{4pt}
\State $w\gets w_0$, $m\gets m_\text{min}$, $F_\text{avg}\gets 0$, $\xi\gets 0$ 
\For{$k=1,\dotsc, K$}
  \LState Draw a mini-batch $\mathcal{B}$ of size $m$
  \LState \raisebox{0pt}[-1pt][-1pt]{$F,g,S\gets$ \Call{Evaluate}{$w$, $\mathcal{B}$}}
  \LState \raisebox{0pt}[-1pt][-1pt]{$\phantom{F,g,S}\mathllap{w}\gets w - \alpha g$} %\Comment{Perform SGD update}
  \LState \raisebox{0pt}[-1pt][-1pt]{$\phantom{F,g,S}\mathllap{\xi}\gets \mu \xi +(1-\mu)\Vert S\Vert_1$}
  \LState \raisebox{0pt}[-1pt][-1pt]{$\phantom{F,g,S}\mathllap{F_\text{avg}}\gets \mu F_\text{avg} + (1-\mu) F$}
  \LState \raisebox{0pt}[-1pt][-1pt]{$\phantom{F,g,S}\mathllap{m}\gets $\Call{round\_\&\_clip}{$\alpha \xi / F_\text{avg}$, $m_\text{min}, m_\text{max}$}}
\EndFor
\end{algorithmic}
Note: \textsc{Evaluate}($w$, $\mathcal{B}$) denotes an evaluation of function value $F(w)$, stochastic gradient $g(w)$ and variance estimate $S(w)$ (Eq.~\ref{eq:sample_variance}) using mini-batch $\mathcal{B}$.
\textsc{round\_\&\_clip}($m$, $m_\text{min}$, $m_\text{max}$) rounds $m$ to the nearest integer and clips it at the provided minimal and maximal values.
\end{algorithm}

\subsection{VARIANCE ESTIMATE}
\label{variance_in_practice}

If the individual gradients $\nabla \ell_i$ in the mini-batch are accessible, then $S$ can be computed directly by Eq.~\eqref{eq:sample_variance}, adding only the computational cost of squaring and summing the gradients. Unfortunately, these individual gradients are not available in practical implementations of the backpropagation algorithm \citep{Rumelhart1986} used to compute gradients in the training of neural networks. A complete discussion of this technical issue is beyond the scope of this paper, but we briefly sketch a solution.

Consider a fully-connected layer in a neural network with weight matrix $W_{l+1}\in\mathbb{R}^{n_l \times n_{l+1}}$.
During the forward pass, the matrix of activations $A_l \in\mathbb{R}^{m\times n_l}$ (containing the activations for each of the $m$ input training samples) is propagated forward by a matrix multiplication,
\begin{equation}
Z_{l+1} = A_l W_{l+1} \in\mathbb{R}^{m\times n_{l+1}}.
\end{equation}
Once the backward pass arrives at this layer, the gradient with respect to $W_{l+1}$ is computed as
\begin{equation}
\label{eq:backprop_matmul}
d W_{l+1} =  A_l^T   d Z_{l+1}.
\end{equation}
The aggregation of individual gradients is \emph{implicit} in this matrix multiplication.
Practical implementations rely on the efficiency of these matrix operations and, even more importantly, it is infeasible to store $m$ individual gradients in memory if the number of parameters $d$ is high.

However, one can similarly compute the \emph{second moment} of the gradients, $\frac{1}{m} \sum_i \left( \nabla \ell_i \right)^{.2}$, that is needed in \eqref{eq:sample_variance} without giving up efficient batch processing.
It is straight-forward to verify that this second moment of gradients with respect to $W^{(l+1)}$ can be computed as
\begin{equation}
\label{eq:backprop_matmul_squared}
\left( A_l^T \right)^{.2} \left( d Z_{l+1} \right)^{.2}.
\end{equation}
In this form, the computation of the gradient variance adds non-negligible but manageable computational cost.
Since it duplicates half of the operations in the backward pass, the additional cost can be pinned down to roughly 25\%.
This is primarily an implementation issue; the cost could be reduced by implementing special matrix operations to compute \eqref{eq:backprop_matmul} and \eqref{eq:backprop_matmul_squared} jointly.

%CABS requires little additional memory, since all maintained quantities are scalar.
%While the gradient variance $S$ has the same dimension as the gradient itself, we only need $\Vert S\Vert_1$, the sum of individual variances.
%This allows us to sum up the variance for each variable (weight matrix, bias variable, convolution filter) immediately after it is computed and free up the memory.

\section{EXPERIMENTS}
\label{experiments}

We evaluate the proposed batch size adaptation method by training convolutional neural networks (CNNs) on four popular image classification benchmark data sets: \textsc{mnist} \citep{LeCun1998}, Street View House Numbers (\textsc{svhn}) \citep{Netzer2011}, as well as \textsc{cifar-10} and \textsc{cifar-100} \citep{Krizhevsky2009}.
While these are small to medium-scale problems by contemporary standards, they exhibit many of the typical difficulties of neural network training.
We opted for these benchmarks to keep the computational cost for a \emph{thorough} evaluation of the method manageable (this required approximately 60 training runs per benchmark, see the following section).

\subsection{EXPERIMENT DESIGN}

We compare against constant batch sizes 16, 32, 64, 128, 256 and 512. To keep the plots readable, we only report results for batch sizes 32, 128 and 512 in the main text; results for the other batch sizes can be found in the supplements.
We also compare against a batch size adaptation based on the criterion \eqref{eq:noisy_grad_norm_rule} used in \cite{Byrd2012} and \cite{De2016}.
Since implementation details differ between these two works, and both combine batch size adaptation with other measures (Newton-CG method in \cite{Byrd2012} and a backtracking line search in \cite{De2016}), we resort to a custom implementation of said criterion.
For a fair comparison, we realize it in a similar manner as \textsc{cabs}.
That is, we use criterion \eqref{eq:noisy_grad_norm_rule}, while keeping the predictive update mechanism for the batch size, the smoothing via exponential moving averages, rounding and clipping exactly as in our \textsc{cabs} implementation described in \textsection\ref{CABS_in_practice} and Algorithm \ref{alg:CABS}.
This method will simply be referred to as \emph{Competitor} in the remainder of this section.

During the optimization process, we periodically evaluate the training loss as well as the classification accuracy on a held-out test set.
Since each method uses a different batch size, both quantities are tracked as a function of the number of accessed training examples, instead of the number of optimization steps. This measure is proportional to wall-clock time up to per-batch overheads that depend on the specific problem and implementation.

The (constant) learning rate for each batch size method was tuned for maximum test accuracy given the fixed budget of accessed training examples.
We tried six candidates $\alpha \in \{0.3, 0.1, 0.06, 0.03, 0.01, 0.006\}$; this relevant range has been determined with a few exploratory experiments.
In addition to the learning rate, the competitor method has a free parameter $\theta$. \cite{De2016} suggest setting it to 1.0, the highest possible noise tolerance, by default.
In our experiments, we found the performance of the method to be fairly sensitive to the choice of $\theta$.
We thus tried $\theta\in \{0.6, 0.8, 1.0\}$ and report results for the best-performing choice.
For \textsc{cabs}, there is no analogous parameter to tune.

\begin{figure}[t]
\centering
\scriptsize
\input{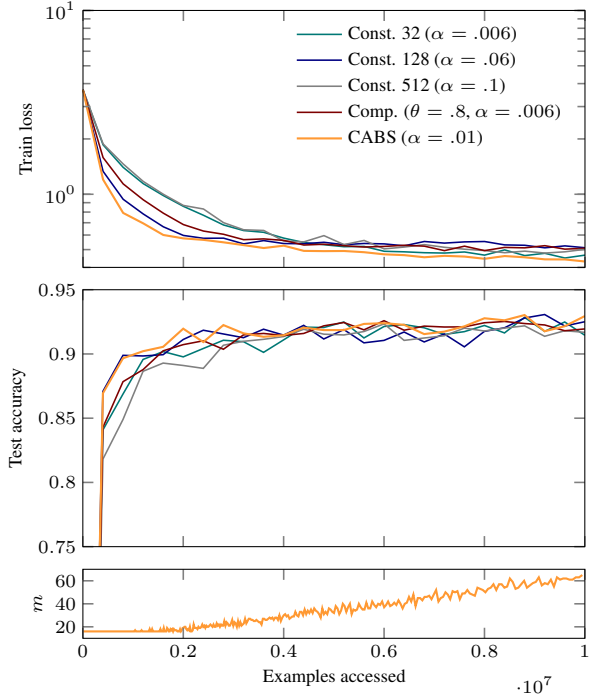}
\caption{Results for \textsc{svhn}. Shared horizontal axis indicates the number of examples used for training. Top and middle panel depict evolution of training loss and test accuracy, respectively, color-coded for different batch size methods, each with its optimal learning rate. Bottom panel shows batch size chosen by \textsc{cabs}.}
\label{fig:results_svhn}
\end{figure}

\begin{figure}[t]
\centering
\scriptsize
\input{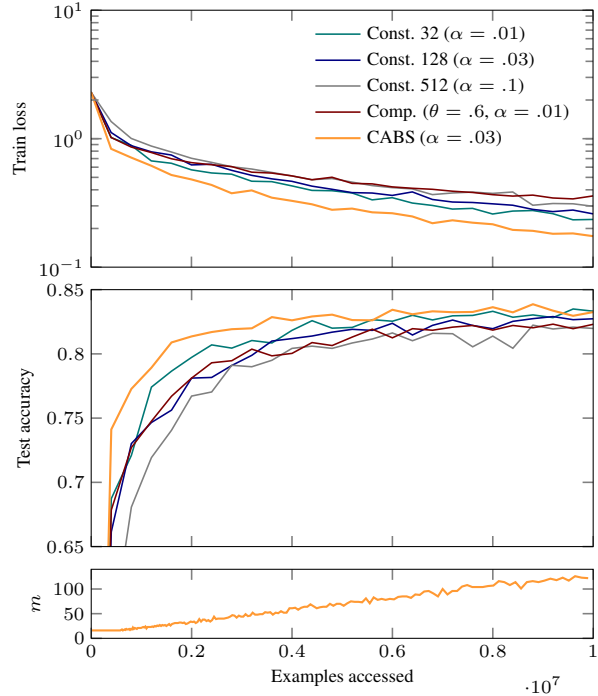}
\caption{Results for \textsc{cifar-10}. Set-up as in Fig.~\ref{fig:results_svhn}.}
\label{fig:results_cifar10}
\end{figure}

\paragraph{MNIST}
We start with experiments on the well-known \textsc{mnist} image classification task of identifying handwritten digits in 28$\times$28 pixel gray scale images.
Our network has two convolutional layers with 5$\times$5 filters (32 and 64 filters, respectively) and subsequent max-pooling over 2$\times$2 windows.
This is followed by a fully-connected layer with 1024 units. 
The activation function is ReLU for all layers.
The output layer has 10 units with softmax activation and we use cross-entropy loss.

\paragraph{SVHN}

Next, we train a CNN on the digit classification task of the Street View House Numbers (\textsc{svhn}) data set. While the task is similar to \textsc{mnist}, the images are in RGB and larger (32$\times$32).
They exhibit real-world views of digits in house numbers, partially with clutter, misalignment and distracting digits at the sides.
We train a CNN with two convolutional layers, each with 64 filters of size 5$\times$5 and subsequent max-pooling over 3$\times$3 windows with stride 2.
They are followed by two fully-connected layers with 256 and 128 units, respectively.
The activation function is ReLU for all layers.
The output layer has 10 units with softmax activation and we use cross-entropy loss.
We apply $L_2$-regularization and perform data augmentation operations (random cropping of 24$\times$24  pixel subimages, left-right mirroring, color distortion) on the training inputs.

\paragraph{CIFAR-10 and CIFAR-100}
Finally, we train CNNs on the \textsc{cifar-10} and \textsc{cifar-100} data sets, where the task is to classify 32$\times$32 pixel RGB images into one of 10 and 100 object categories, respectively.
For \textsc{cifar-10}, we crop the images to 24$\times$24 pixels and train a CNN with two convolutional layers, each with 64 filters of size 5x5 and subsequent max-pooling over 3x3 windows with stride 2.
They are followed by two fully-connected layers with 384 and 192 units, respectively.
The activation function is ReLU for all layers.
The output layer has 10 units with softmax activation and we use cross-entropy loss.
We perform data augmentation operations (random cropping, left-right mirroring, color distortion) on the training set.

For \textsc{cifar-10}0, we use the full 32$\times$32 image and add a third convolutional layer (64 filters of size $5 \times 5$ followed by max pooling). The fully-connected layers have 512 and 256 units, respectively, and the output layer has 100 units. We add $L_2$-regularization.

\subsection{RESULTS AND DISCUSSION}

\begin{figure}[t]
\centering
\scriptsize
% This file was created by matplotlib2tikz v0.6.3.
\begin{tikzpicture}

\definecolor{color1}{rgb}{1,0.6,0.2}
\definecolor{color0}{rgb}{0,0.4717,0.4604}

\begin{groupplot}[group style={group size=1 by 3, vertical sep=.3cm}]
\nextgroupplot[
ylabel={Train loss},
xmin=0, xmax=8000000,
ymin=0.9, ymax=10,
ymode=log,
axis on top,
width=\figurewidth,
height=\figureheight,
xtick scale label code/.code={},
xticklabels={},
mystyle,
legend entries={{Const. 32 ($\alpha=.01$)},{Const. 128 ($\alpha=.03$)},{Const. 512 ($\alpha=.06$)},{Comp. ($\theta=.6$, $\alpha=.006$)},{CABS ($\alpha=.01$)}},
legend cell align={left}
]
\addplot [semithick, color0]
table {%
0 5.6335257490476
400000 3.35691821651581
800000 2.78699110219112
1200000 2.48257953234208
1600000 2.23883897409989
2000000 2.26461899739045
2400000 2.15978199319962
2800000 2.09708820894743
3200000 1.97076720037521
3600000 1.95466169409263
4000000 1.91988318852889
4400000 1.90650653648071
4800000 1.91273740010384
5200000 1.82722252263473
5600000 1.84551184414289
6000000 1.82571102105654
6400000 1.84114421139925
6800000 1.84119209952843
7200000 1.80328029203109
7600000 1.74979790204611
8000000 1.75284341894663
};
\addplot [semithick, blue!50.899999999999999!black]
table {%
0 5.63546615380507
400000 3.66864059215937
800000 2.96933091603793
1200000 2.60169655848772
1600000 2.38111215371352
2000000 2.28560045437935
2400000 2.15870755910873
2800000 2.07999242116243
3200000 2.06238332467201
3600000 1.86584600271323
4000000 1.85959869470352
4400000 1.85891774220344
4800000 1.82300986540623
5200000 1.79390866022844
5600000 1.84303604945158
6000000 1.79550737753893
6400000 1.72245354682971
6800000 1.78934750801478
7200000 1.77451584125176
7600000 1.76894748363739
8000000 1.69420278530854
};
\addplot [semithick, gray]
table {%
0 5.63640586953414
400384 4.11740423503675
800256 3.75484757674368
1200128 3.31134866413317
1600000 3.01615495430796
2000384 2.79207484345687
2400256 2.7208480583994
2800128 2.51681495967664
3200000 2.34686265493694
3600384 2.31586153883683
4000256 2.28632494022972
4400128 2.10229951456973
4800000 2.0354775755029
5200384 1.96535374616322
5600256 1.95255708067041
6000128 2.05598028082597
6400000 2.14744658219187
6800384 1.82919539275922
7200256 1.84926655417994
7600128 1.82746094151547
8000000 1.88980618276094
};
\addplot [semithick, red!49.059999999999995!black]
table {%
0 5.63241353401771
400015 3.9005544980367
800026 3.22168581302349
1200015 2.873995640339
1600018 2.68595157525478
2000006 2.5493682775742
2400016 2.44346023828555
2800009 2.3320144812266
3200002 2.27180104377942
3600002 2.1576993557123
4000003 2.14611089229584
4400004 2.08103056443043
4800022 2.09922982179202
5200017 2.04426898711767
5600007 2.01722440047142
6000025 1.96637512476016
6400032 2.0755432783029
6800016 1.99708028940054
7200021 1.99498580357967
7600008 1.88937174356901
8000016 1.92855727672577
};
\addplot [thick, color1]
table {%
0 5.63318135188176
400000 2.98240736814646
800000 2.51743885187002
1200000 2.38122213192475
1600000 2.20234073736729
2000002 2.27656082006601
2400002 2.12121861408918
2800031 1.83001860594138
3200031 1.73760526302533
3600034 1.60366198955438
4000005 1.56674599647522
4400087 1.50409081043341
4800036 1.41503045497796
5200017 1.3838501068262
5600006 1.32600557804108
6000130 1.33737110785949
6400103 1.2801290842203
6800115 1.25430098863748
7200157 1.24214646449456
7600112 1.20820969190353
8000043 1.22744097770789
};
\nextgroupplot[
ylabel={Test accuracy},
xmin=0, xmax=8000000,
ymin=0.25, ymax=0.55,
axis on top,
width=\figurewidth,
height=\figureheight,
xtick scale label code/.code={},
xticklabels={},
mystyle
]
\addplot [semithick, color0]
table {%
0 0.0154246794871795
400000 0.310697115384615
800000 0.378205128205128
1200000 0.418770032051282
1600000 0.447015224358974
2000000 0.436899038461538
2400000 0.437900641025641
2800000 0.431790865384615
3200000 0.467748397435897
3600000 0.452524038461538
4000000 0.459735576923077
4400000 0.468349358974359
4800000 0.463141025641026
5200000 0.463341346153846
5600000 0.468349358974359
6000000 0.462439903846154
6400000 0.456630608974359
6800000 0.465144230769231
7200000 0.453125
7600000 0.468249198717949
8000000 0.464242788461538
};
\addplot [semithick, blue!50.899999999999999!black]
table {%
0 0.00701121794871795
400000 0.251502403846154
800000 0.367588141025641
1200000 0.400040064102564
1600000 0.430689102564103
2000000 0.420472756410256
2400000 0.439903846153846
2800000 0.43359375
3200000 0.444310897435897
3600000 0.457932692307692
4000000 0.449919871794872
4400000 0.461137820512821
4800000 0.456129807692308
5200000 0.455228365384615
5600000 0.454527243589744
6000000 0.458934294871795
6400000 0.460236378205128
6800000 0.452123397435897
7200000 0.442808493589744
7600000 0.450220352564103
8000000 0.450520833333333
};
\addplot [semithick, gray]
table {%
0 0.0105879934210526
400384 0.207956414473684
800256 0.253289473684211
1200128 0.309621710526316
1600000 0.345189144736842
2000384 0.382606907894737
2400256 0.394325657894737
2800128 0.391652960526316
3200000 0.421052631578947
3600384 0.413342927631579
4000256 0.410670230263158
4400128 0.437088815789474
4800000 0.439967105263158
5200384 0.436472039473684
5600256 0.425986842105263
6000128 0.426295230263158
6400000 0.402652138157895
6800384 0.440583881578947
7200256 0.423108552631579
7600128 0.435135690789474
8000000 0.419819078947368
};
\addplot [semithick, red!49.059999999999995!black]
table {%
0 0.0118189102564103
400015 0.229867788461538
800026 0.340344551282051
1200015 0.373497596153846
1600018 0.400140224358974
2000006 0.412660256410256
2400016 0.427784455128205
2800009 0.430989583333333
3200002 0.440805288461538
3600002 0.446314102564103
4000003 0.452223557692308
4400004 0.456630608974359
4800022 0.461338141025641
5200017 0.461538461538462
5600007 0.459435096153846
6000025 0.470152243589744
6400032 0.452524038461538
6800016 0.468149038461538
7200021 0.454527243589744
7600008 0.477764423076923
8000016 0.475861378205128
};
\addplot [thick, color1]
table {%
0 0.0107171474358974
400000 0.348157051282051
800000 0.414663461538462
1200000 0.426983173076923
1600000 0.457231570512821
2000002 0.446915064102564
2400002 0.461838942307692
2800031 0.487079326923077
3200031 0.483974358974359
3600034 0.490785256410256
4000005 0.482271634615385
4400087 0.484875801282051
4800036 0.48838141025641
5200017 0.489383012820513
5600006 0.49599358974359
6000130 0.489282852564103
6400103 0.496594551282051
6800115 0.487379807692308
7200157 0.490785256410256
7600112 0.492788461538462
8000043 0.48828125
};
\nextgroupplot[
xlabel={Examples accessed},
ylabel={$m$},
xmin=0, xmax=8000000,
ymin=0, ymax=200,
axis on top,
width=\figurewidth,
height=.5\figureheight,
mystyle
]
\addplot [thick, color1]
table {%
0 16
16000 16
32000 16
48000 16
64000 16
80000 16
96000 16
112000 16
128000 16
144000 16
160000 16
176000 16
192000 16
208000 16
224000 16
240000 16
256000 16
272000 16
288000 16
304000 16
320000 16
336000 16
352000 16
368000 16
384000 16
400000 16
416000 16
432000 16
448000 16
464000 16
480000 16
496000 16
512000 16
528000 16
544000 16
560000 16
576000 16
592000 16
608000 16
624000 16
640000 16
656000 16
672000 16
688000 16
704000 16
720000 16
736000 16
752000 16
768000 16
784000 16
800000 16
816000 16
832000 16
848000 16
864000 16
880000 16
896000 16
912000 16
928000 16
944000 16
960000 16
976000 16
992000 16
1008000 16
1024000 16
1040000 16
1056000 16
1072000 16
1088000 16
1104000 16
1120000 16
1136000 16
1152000 16
1168000 16
1184000 16
1200000 16
1216000 16
1232000 16
1248000 16
1264000 16
1280000 16
1296000 16
1312000 16
1328000 16
1344000 16
1360000 16
1376000 16
1392000 16
1408000 16
1424000 16
1440000 16
1456000 16
1472000 16
1488000 16
1504000 16
1520000 16
1536000 16
1552000 16
1568000 16
1584000 16
1600000 16
1616000 16
1632000 16
1648000 16
1664000 16
1680000 16
1696000 16
1712000 16
1728002 16
1744002 16
1760007 16
1776016 16
1792016 16
1808039 16
1824057 16
1840114 16
1856123 16
1872177 16
1888206 16
1904214 16
1920454 16
1936741 16
1952776 16
1969113 17
1985376 16
2001500 17
2018477 17
2035023 16
2051229 16
2068306 17
2084934 17
2101590 17
2119798 19
2137647 18
2154805 18
2172877 19
2190918 18
2209182 19
2228481 19
2247210 18
2266633 20
2286326 19
2305824 18
2326698 21
2346808 20
2367750 20
2389447 22
2411593 23
2434777 23
2457190 23
2480726 24
2504400 24
2529849 25
2554620 26
2581000 27
2607064 28
2634355 27
2662139 28
2690717 27
2720225 31
2750395 30
2781644 32
2813752 33
2846595 32
2880824 33
2915338 35
2951134 36
2987980 39
3026640 38
3065593 39
3105996 41
3147689 41
3191428 44
3236059 45
3281854 45
3329632 49
3378157 48
3429041 54
3480858 53
3536048 57
3592761 58
3651128 59
3711745 63
3774847 67
3840676 64
3908676 69
3978951 72
4052257 74
4128476 76
4206779 80
4288309 82
4373244 86
4461145 89
4551483 90
4646075 94
4743590 98
4844523 102
4949453 106
5056804 111
5169178 117
5284693 121
5404898 123
5527807 131
5656168 128
5787546 132
5923622 136
6064018 145
6209440 148
6358603 146
6512684 153
6669932 161
6831424 165
6998540 172
7170195 174
7346288 179
7527248 181
7712994 183
7903256 195
};
\end{groupplot}

\end{tikzpicture}
\caption{Results for \textsc{cifar-100}. Set-up as in Fig.~\ref{fig:results_svhn}.}
\label{fig:results_cifar100}
\end{figure}
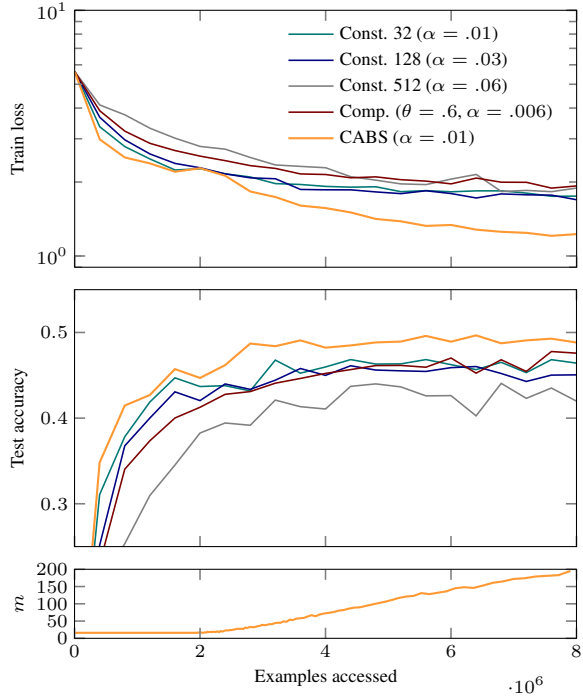

\begin{figure}[t]
\centering
\scriptsize
\input{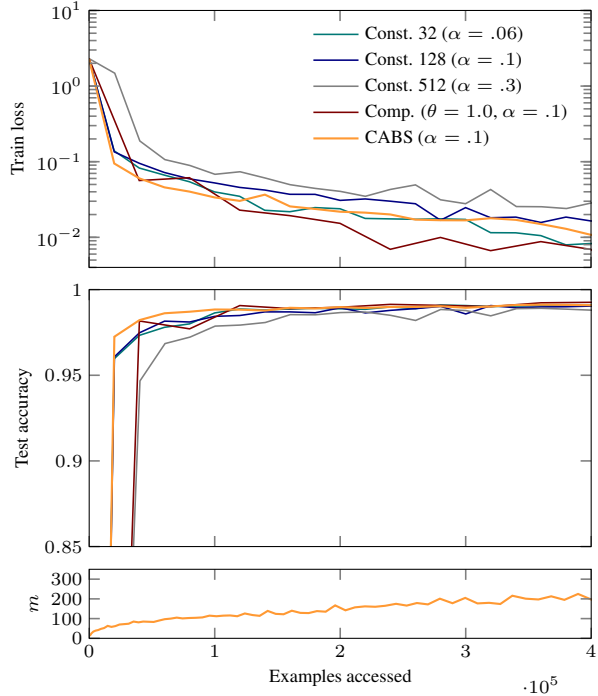}
\vspace{-8pt}
\caption{Results for \textsc{mnist}. Set-up as in Fig.~\ref{fig:results_svhn}.}
\label{fig:results_mnist}
\end{figure}

On \textsc{svhn}, \textsc{cifar-10} and \textsc{cifar-100} (Figures \ref{fig:results_svhn}, \ref{fig:results_cifar10} and \ref{fig:results_cifar100}), \textsc{cabs} yields significantly faster decrease in training loss with the curve contiuously lying below all others. It also achieves the best test set accuracy of all methods on all three problems.
While the margin over the second-best method is very small on \textsc{svhn}, it amounts to a noticeable 0.4 percentage points on \textsc{cifar-10} and even 1.4 points on \textsc{cifar-100}.

Surprisingly, on \textsc{mnist}---the \emph{least} complex of the benchmark problems we investigated---our method is outperformed by the small constant batch size of 32 and the competitor method, which also chooses very small batch sizes throughout.
Our method does, however, surpass non-adaptive larger batch sizes (128, 512) in terms of speed and all contenders reach virtually the same test set accuracy on this problem.
\textsc{cabs} makes rapid progress initially, but seems to choose unnecessarily large batch sizes later on. The resulting high per-iteration cost evidently can not be compensated by the higher learning rate it enables (.1 for \textsc{cabs} compared to .01 with constant batch size of 32).
We conjecture that \textsc{cabs} overestimates the gradient variance due to the homogeneous structure of the \textsc{mnist} data set; if the distribution of gradients is very closely-centered, outliers in a few coordinate directions lead to comparably high variance estimates.

Overall, \textsc{cabs} outperforms alternative batch size schemes on three out of the four benchmark problems we investigated and the benefits seem to increase with the complexity of the problem (\textsc{mnist} $\rightarrow$ \textsc{svhn} $\rightarrow$ \textsc{cifar-10} $\rightarrow$ \textsc{cifar-100}). When considering the \textsc{cabs} batch size schedules, depicted in the bottom panels of Figures \ref{fig:results_svhn} to \ref{fig:results_mnist}, a common behavior (with the exception of \textsc{mnist}) seems to be that \textsc{cabs} uses the minimal batch size (16 in our experiments) for a considerable portion of the training process and increases approximately linearly afterwards.

\begin{figure}[t]
\centering
\scriptsize
\input{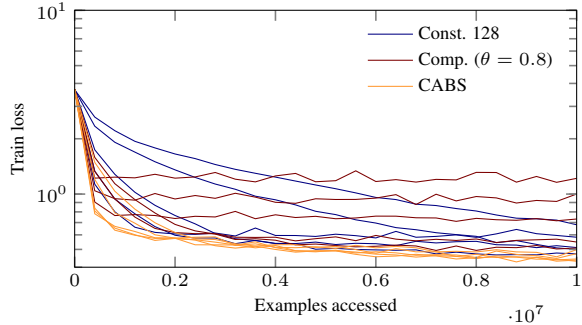}
\caption{Learning rate sensitivity on \textsc{svhn}. Families of training loss curves for \textsc{cabs}, Competitor and a constant batch size (color-coded). Each individual curve corresponds to a learning rate $\alpha \in \{.1, .06, .03, .01, .006\}$.}
\label{fig:lr_sensitivity_svhn}
\end{figure}

\begin{figure}[t]
\centering
\scriptsize
\input{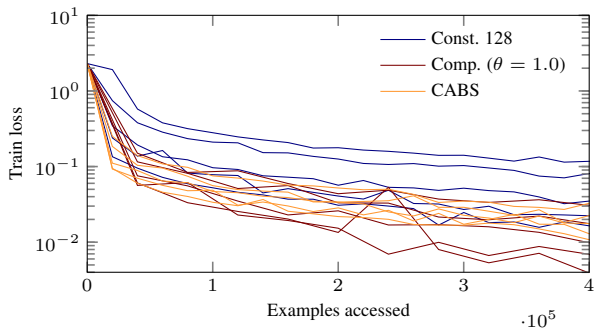}
\caption{Learning rate sensitivity on \textsc{mnist}. Set-up as in Fig.~\ref{fig:lr_sensitivity_svhn}.}
\label{fig:lr_sensitivity_mnist}
\end{figure}

Finally, we present our findings regarding the sensitivity to the choice of learning rate when using \textsc{cabs}.
As detailed above, the coupling of learning rate and batch size in \textsc{cabs} can be seen as tayloring the noise level to the chosen learning rate.
This suggests that the performance of the optimizer should be less sensitive to the choice of learning rate when adapting the batch size with our method.
Indeed, this became evident in our experiments by considering the families of training loss curves for various learning rates. 
Figures \ref{fig:lr_sensitivity_svhn} and \ref{fig:lr_sensitivity_mnist} compare a constant batch size, the competitor method and \textsc{cabs} on the \textsc{svhn} and \textsc{mnist} benchmark problems, respectively.
\textsc{cabs} significantly reduces the dependency of the performance on the learning rate compared to both the constant batch size and the competing adaptive method.
In practical applications, this finding could drastically simplify the often tedious process of learning rate tuning.

\section{CONCLUSION}

We proposed \textsc{cabs}, a practical rule for dynamic batch size adaptation based on estimates of the gradient variance and coupled to the chosen learning rate as well as optimization progress represented by the function value.
In our experiments, \textsc{cabs} was able to speed up \textsc{sgd} training in neural networks and simplify the tuning of the learning rate.
In contrast to existing methods, it does not introduce any additional free parameters.
A TensorFlow\footnote{\url{http://tensorflow.org}} implementation of \textsc{sgd} with \textsc{cabs} can be found on \url{http://github.com/ProbabilisticNumerics/cabs}.

\appendix

\section{MATHEMATICAL DETAILS}

\label{appendix_mathematical_details}

\paragraph{Proof of Equation (\ref{eq:descent_direction})} By the Cauchy-Schwarz inequality, $\langle g, \nabla F\rangle = \langle g, g \rangle - \langle g, g-\nabla F\rangle  \geq \Vert g \Vert^2 - \Vert g\Vert \Vert g-\nabla F\Vert  = \Vert g \Vert (\Vert g\Vert - \Vert g-\nabla F\Vert)$. This becomes positive if $\Vert g - \nabla F\Vert < \Vert g \Vert$.

\paragraph{Solving the Maximization Problem (\ref{eq:cost_sensitive_gain_maximization})}
We want to maximize
\begin{equation}
U(m) = \frac{\mathbf{E}[\mathcal{G}]}{m} = \frac{2 \alpha - L \alpha^2}{2m} \Vert \nabla F\Vert^2  - \frac{L \alpha^2}{2m^2} \text{tr}(\Sigma).
\end{equation}
Setting the derivative
\begin{equation}
U^\prime(m) = - \frac{2 \alpha - L \alpha^2}{2m^2} \Vert \nabla F\Vert^2 + \frac{L \alpha^2}{m^3} \text{tr}(\Sigma)
\end{equation}
to zero and rearranging yields \eqref{eq:lipschitz_bound_rule}.

\paragraph{Proof of Equation (\ref{eq:grad_norm_to_distance_to_optimality_convexity})}
The definition of strong convexity is that, for all $w,u\in\mathbb{R}^d$
\begin{equation}
\label{eq:strong_convexity}
F(u) \geq F(w) + \nabla F(w)^T(u-w) + \frac{\mu}{2} \Vert u-w\Vert^2
\end{equation}
for $\mu >0$.
From there, the proof is identical to that of Equation \eqref{eq:grad_norm_to_distance_to_optimality} in the main text.
We minimize both sides of the inequality. The left-hand side has minimal value $F_\ast$.
The gradient with respect to $u$ of the right-hand side is $\nabla F(w) + \mu (u-w)$.
By setting this to zero we find the minimizer $u=-\nabla F(w)/\mu + w$.
Inserting this back into \eqref{eq:strong_convexity} and rearranging yields \eqref{eq:grad_norm_to_distance_to_optimality_convexity}.\hfill$\square$

%\subsubsection*{Acknowledgements}
%Use unnumbered third level headings for the acknowledgements title.
%All acknowledgements go at the end of the paper.

\subsubsection*{References}
{\def\section*#1{} % teporarilly suppres section headings
\bibliographystyle{icml2017}
\bibliography{references.bib}
}

\clearpage
\section*{Supplementary Material}

\begin{figure}[h]
\centering
\scriptsize
\input{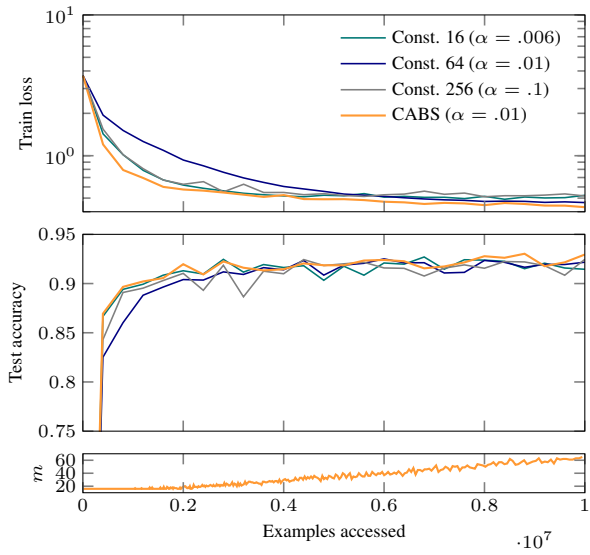}
\caption{Additional results for \textsc{svhn}.}
\label{fig:results_svhn_supp}
\end{figure}

\begin{figure}[h]
\centering
\scriptsize
\input{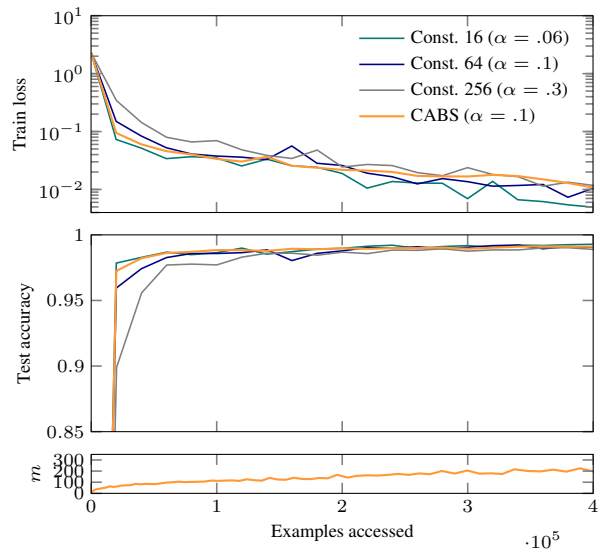}
\caption{Additional results for \textsc{mnist}.}
\label{fig:results_mnist_supp}
\end{figure}

Figures \ref{fig:results_svhn_supp} to \ref{fig:results_mnist_supp} show results for constant batch sizes 16, 64 and 256, which have been omitted from the main text for lack of space.

\setlength\figureheight{4.2cm}
\setlength\floatsep{0pt}

\vfill

\begin{figure}[h]
\centering
\scriptsize
% This file was created by matplotlib2tikz v0.6.3.
\begin{tikzpicture}

\definecolor{color1}{rgb}{1,0.6,0.2}
\definecolor{color0}{rgb}{0,0.4717,0.4604}

\begin{groupplot}[group style={group size=1 by 3, vertical sep=.3cm}]
\nextgroupplot[
ylabel={Train loss},
xmin=0, xmax=10000000,
ymin=0.1, ymax=10,
ymode=log,
axis on top,
width=\figurewidth,
height=\figureheight,
xtick scale label code/.code={},
xticklabels={},
mystyle,
legend cell align={left},
legend entries={{Const. 16 ($\alpha=.006$)},{Const. 64 ($\alpha=.01$)},{Const. 256 ($\alpha=.06$)},{CABS ($\alpha=.03$)}}
]
\addplot [semithick, color0]
table {%
0 2.30362619094849
400000 0.986136943936348
800000 0.823385021972656
1200000 0.706046251583099
1600000 0.64039548163414
2000000 0.567263424444199
2400000 0.531607158100605
2800000 0.501964652955532
3200000 0.445617201918364
3600000 0.482171088516712
4000000 0.398421789225936
4400000 0.395592461612821
4800000 0.363303778798878
5200000 0.353360690587759
5600000 0.359829170304537
6000000 0.309036552500725
6400000 0.308250676262379
6800000 0.291445182660222
7200000 0.276480584087968
7600000 0.27232091396749
8000000 0.281168224471808
8400000 0.267029322962463
8800000 0.256666978701204
9200000 0.233562242624909
9600000 0.231735528430715
10000000 0.223295853217691
};
\addplot [semithick, blue!50.899999999999999!black]
table {%
0 2.302995518232
400000 1.15223794258558
800000 0.947206492989491
1200000 0.843914553905145
1600000 0.813307871421178
2000000 0.723661316319918
2400000 0.656712103539552
2800000 0.615360318468167
3200000 0.621313196344253
3600000 0.561490235229333
4000000 0.526176026998422
4400000 0.504745757159514
4800000 0.475401558937171
5200000 0.467214093567469
5600000 0.439993783545036
6000000 0.422664260443969
6400000 0.419675168223106
6800000 0.406427739235835
7200000 0.3831175855146
7600000 0.373250315777766
8000000 0.352782839670395
8400000 0.337726645171642
8800000 0.356704619928048
9200000 0.355053938256624
9600000 0.346841281900803
10000000 0.293760608595151
};
\addplot [semithick, gray]
table {%
0 2.30331480808747
400128 1.15417546798021
800000 0.956174498949295
1200128 0.79122584599715
1600000 0.73166122344824
2000128 0.659807082934257
2400000 0.618524572023979
2800128 0.560159480724579
3200000 0.569176631096082
3600128 0.512726857112004
4000000 0.486968968923275
4400128 0.456699988016715
4800000 0.433644070839271
5200128 0.425562989253264
5600000 0.432780122145628
6000128 0.396560672001961
6400000 0.392881565369093
6800128 0.344730844864478
7200000 0.334200768898695
7600128 0.333437035481135
8000000 0.309428721666336
8400128 0.324939563106268
8800000 0.310305921695171
9200128 0.285290307341478
9600000 0.284331622796181
10000128 0.289674501388501
};
\addplot [thick, color1]
table {%
0 2.3017444060399
400012 0.83372778006089
800010 0.712748629924579
1200007 0.615703622500102
1600030 0.521758437156677
2000003 0.48184790290319
2400025 0.436486729444602
2800038 0.37607419414398
3200023 0.395877710519693
3600021 0.347484793418493
4000027 0.328294663857191
4400048 0.308613022550558
4800049 0.279567014712554
5200018 0.285732850432396
5600042 0.267226671561217
6000028 0.262846105755904
6400027 0.24816377269916
6800086 0.21977882201855
7200014 0.231604602856514
7600094 0.22189734914364
8000001 0.215740362038979
8400041 0.195067944626013
8800084 0.191746024749218
9200024 0.181887858189069
9600005 0.183089951292062
10000006 0.174282897741367
};
\nextgroupplot[
ylabel={Test accuracy},
xmin=0, xmax=10000000,
ymin=0.65, ymax=0.85,
axis on top,
width=\figurewidth,
height=\figureheight,
xtick scale label code/.code={},
xticklabels={},
mystyle
]
\addplot [semithick, color0]
table {%
0 0.0997
400000 0.6894
800000 0.7342
1200000 0.7671
1600000 0.7883
2000000 0.7938
2400000 0.8076
2800000 0.8077
3200000 0.8164
3600000 0.81
4000000 0.8224
4400000 0.8161
4800000 0.8223
5200000 0.8243
5600000 0.8265
6000000 0.8299
6400000 0.8249
6800000 0.8319
7200000 0.8258
7600000 0.8204
8000000 0.8236
8400000 0.8264
8800000 0.8293
9200000 0.8329
9600000 0.8317
10000000 0.8309
};
\addplot [semithick, blue!50.899999999999999!black]
table {%
0 0.0960536858974359
400000 0.635717147435897
800000 0.705829326923077
1200000 0.738581730769231
1600000 0.742688301282051
2000000 0.76582532051282
2400000 0.784354967948718
2800000 0.783553685897436
3200000 0.786358173076923
3600000 0.795472756410256
4000000 0.80488782051282
4400000 0.805789262820513
4800000 0.809595352564103
5200000 0.813802083333333
5600000 0.818709935897436
6000000 0.817708333333333
6400000 0.819711538461538
6800000 0.822716346153846
7200000 0.824719551282051
7600000 0.82401842948718
8000000 0.824719551282051
8400000 0.825220352564103
8800000 0.821814903846154
9200000 0.820913461538462
9600000 0.818609775641026
10000000 0.831630608974359
};
\addplot [semithick, gray]
table {%
0 0.0872395833333333
400128 0.62510016025641
800000 0.703125
1200128 0.750701121794872
1600000 0.762419871794872
2000128 0.776943108974359
2400000 0.785857371794872
2800128 0.798377403846154
3200000 0.785657051282051
3600128 0.799879807692308
4000000 0.802884615384615
4400128 0.81229967948718
4800000 0.819611378205128
5200128 0.813000801282051
5600000 0.811999198717949
6000128 0.821213942307692
6400000 0.817808493589744
6800128 0.821814903846154
7200000 0.819210737179487
7600128 0.819511217948718
8000000 0.822215544871795
8400128 0.819210737179487
8800000 0.822816506410256
9200128 0.823317307692308
9600000 0.822315705128205
10000128 0.822816506410256
};
\addplot [thick, color1]
table {%
0 0.100961538461538
400012 0.741185897435897
800010 0.772736378205128
1200007 0.78926282051282
1600030 0.808894230769231
2000003 0.813601762820513
2400025 0.817007211538462
2800038 0.819210737179487
3200023 0.819911858974359
3600021 0.828625801282051
4000027 0.826121794871795
4400048 0.829126602564103
4800049 0.830528846153846
5200018 0.826322115384615
5600042 0.826121794871795
6000028 0.834334935897436
6400027 0.830829326923077
6800086 0.833133012820513
7200014 0.832431891025641
7600094 0.832632211538462
8000001 0.836338141025641
8400041 0.832331730769231
8800084 0.838641826923077
9200024 0.833633814102564
9600005 0.829627403846154
10000006 0.832532051282051
};
\nextgroupplot[
xlabel={Examples accessed},
ylabel={$m$},
xmin=0, xmax=10000000,
ymin=0, ymax=140,
axis on top,
width=\figurewidth,
height=.5\figureheight,
mystyle
]
\addplot [thick, color1]
table {%
0 16
16000 16
32000 16
48000 16
64000 16
80000 16
96000 16
112000 16
128000 16
144000 16
160000 16
176000 16
192000 16
208000 16
224000 16
240000 16
256000 16
272000 16
288015 16
304015 16
320015 16
336015 16
352015 16
368046 16
384060 16
400060 16
416071 16
432092 16
448128 16
464295 16
480296 16
496310 16
512580 16
528946 16
545172 16
562075 16
578595 16
595712 17
612622 18
630302 16
647154 17
664304 18
681560 17
699468 19
717277 18
735495 18
753997 19
772728 17
791251 19
810236 18
829846 22
849446 19
869443 21
889067 20
908862 19
928798 21
949482 21
970841 21
992221 21
1013330 22
1035261 23
1056630 22
1078473 24
1100101 20
1123064 23
1145932 23
1168721 23
1192003 23
1214947 24
1239050 23
1264084 25
1288993 24
1314255 26
1339478 25
1365446 27
1390973 25
1417080 27
1443256 26
1470264 27
1497079 25
1524330 27
1552113 29
1580512 26
1608886 29
1637449 30
1666708 31
1696493 31
1726934 31
1757575 32
1788964 32
1820276 31
1852217 32
1884456 29
1916061 30
1948498 32
1981306 34
2015164 33
2048412 34
2081569 31
2116149 33
2152165 39
2187798 34
2223746 38
2259862 38
2297007 35
2333722 37
2371052 34
2408139 37
2446512 44
2485201 39
2524598 40
2563002 40
2602586 40
2643326 40
2683571 41
2724969 42
2766910 44
2810276 47
2853929 46
2896431 42
2940592 46
2985034 44
3029568 44
3074509 49
3121310 45
3168429 46
3215116 52
3264019 48
3313369 48
3363296 49
3413802 49
3464463 53
3515434 52
3566726 53
3618566 48
3672118 52
3725136 51
3780083 55
3835166 52
3891490 52
3948728 61
4005945 61
4062843 62
4122685 58
4182930 62
4243972 64
4305025 64
4366524 69
4429024 65
4492674 61
4557185 63
4621457 64
4687744 64
4755502 70
4823272 66
4890462 70
4960475 70
5029317 66
5100102 71
5173730 75
5246532 74
5321920 74
5396619 73
5472629 81
5548779 71
5626419 83
5706309 78
5785457 80
5865078 78
5947980 80
6031460 79
6115424 84
6200155 85
6284319 85
6369764 82
6455725 89
6544105 88
6633169 95
6723057 96
6816077 91
6907558 85
7000722 100
7096597 89
7192744 96
7288991 96
7387013 105
7489610 108
7590833 104
7694342 104
7798783 104
7904154 106
8011770 107
8119084 116
8228844 114
8340024 108
8450890 114
8562366 101
8677963 114
8794379 116
8909794 114
9027407 117
9147013 121
9269419 118
9392174 123
9513305 117
9638931 126
9769233 123
9899733 122
};
\end{groupplot}

\end{tikzpicture}
\caption{Additional results for \textsc{cifar-10}.}
\label{fig:results_cifar10_supp}
\end{figure}

\newpage
\phantom{.}
\vfill
\begin{figure}[h]
\centering
\scriptsize
% This file was created by matplotlib2tikz v0.6.3.
\begin{tikzpicture}

\definecolor{color1}{rgb}{1,0.6,0.2}
\definecolor{color0}{rgb}{0,0.4717,0.4604}

\begin{groupplot}[group style={group size=1 by 3, vertical sep=.3cm}]
\nextgroupplot[
ylabel={Train loss},
xmin=0, xmax=8000000,
ymin=0.9, ymax=10,
ymode=log,
axis on top,
width=\figurewidth,
height=\figureheight,
xtick scale label code/.code={},
xticklabels={},
legend entries={{Const. 16 ($\alpha=.006$)},{Const. 64 ($\alpha=.01$)},{Const. 256 ($\alpha=.06$)},{CABS ($\alpha=.01$)}},
mystyle,
legend cell align={left}
]
\addplot [semithick, color0]
table {%
0 5.63308108444214
400000 3.18198037338257
800000 2.58377515048981
1200000 2.37392250366211
1600000 2.21458798160553
2000000 2.18560136013031
2400000 2.06206948413849
2800000 2.01097735204697
3200000 1.93517815227509
3600000 1.91941186294556
4000000 1.96253252372742
4400000 1.93399294910431
4800000 1.88326603908539
5200000 1.91053032474518
5600000 1.89278342266083
6000000 1.87089096240997
6400000 1.88914699115753
6800000 1.85519582614899
7200000 1.92192894821167
7600000 1.89264756469727
8000000 1.81295440940857
};
\addplot [semithick, blue!50.899999999999999!black]
table {%
0 5.63422242189065
400000 3.79477903476128
800000 3.17118668556213
1200000 2.79780824826314
1600000 2.56915586728316
2000000 2.35122057413444
2400000 2.25034561447608
2800000 2.09677397517058
3200000 2.0114415983359
3600000 1.98088643107659
4000000 1.87551741187389
4400000 1.81692139689739
4800000 1.81679551265179
5200000 1.75833187577052
5600000 1.67263785386697
6000000 1.64105811485877
6400000 1.68399307666681
6800000 1.60493927659133
7200000 1.61367656099491
7600000 1.57723888143515
8000000 1.55889731874833
};
\addplot [semithick, gray]
table {%
0 5.63285232201601
400128 3.70429437588423
800000 3.17222586656228
1200128 2.94406230633075
1600000 2.6574923931024
2000128 2.34923045451824
2400000 2.26631365678249
2800128 2.20913855234782
3200000 2.08636827346606
3600128 2.04999009156838
4000000 2.02438327899346
4400128 2.1685303969261
4800000 1.96331911820632
5200128 2.01863324030852
5600000 1.8894126751484
6000128 1.95480925303239
6400000 1.83155455344763
6800128 1.86450513815268
7200000 1.74141813852848
7600128 1.8237670415487
8000000 1.81917028243725
};
\addplot [thick, color1]
table {%
0 5.63318135188176
400000 2.98240736814646
800000 2.51743885187002
1200000 2.38122213192475
1600000 2.20234073736729
2000002 2.27656082006601
2400002 2.12121861408918
2800031 1.83001860594138
3200031 1.73760526302533
3600034 1.60366198955438
4000005 1.56674599647522
4400087 1.50409081043341
4800036 1.41503045497796
5200017 1.3838501068262
5600006 1.32600557804108
6000130 1.33737110785949
6400103 1.2801290842203
6800115 1.25430098863748
7200157 1.24214646449456
7600112 1.20820969190353
8000043 1.22744097770789
};
\nextgroupplot[
ylabel={Test accuracy},
xmin=0, xmax=8000000,
ymin=0.25, ymax=0.55,
axis on top,
width=\figurewidth,
height=\figureheight,
xtick scale label code/.code={},
xticklabels={},
mystyle
]
\addplot [semithick, color0]
table {%
0 0.0103
400000 0.3377
800000 0.4151
1200000 0.4366
1600000 0.461
2000000 0.4429
2400000 0.4499
2800000 0.4595
3200000 0.4655
3600000 0.4672
4000000 0.4638
4400000 0.4658
4800000 0.4699
5200000 0.4644
5600000 0.4596
6000000 0.4618
6400000 0.4639
6800000 0.46
7200000 0.458
7600000 0.457
8000000 0.461
};
\addplot [semithick, blue!50.899999999999999!black]
table {%
0 0.0111177884615385
400000 0.260316506410256
800000 0.344551282051282
1200000 0.394330929487179
1600000 0.419671474358974
2000000 0.434895833333333
2400000 0.453525641025641
2800000 0.453125
3200000 0.453125
3600000 0.468249198717949
4000000 0.455929487179487
4400000 0.462139423076923
4800000 0.456630608974359
5200000 0.457632211538462
5600000 0.47275641025641
6000000 0.466145833333333
6400000 0.457732371794872
6800000 0.461838942307692
7200000 0.461838942307692
7600000 0.465645032051282
8000000 0.46484375
};
\addplot [semithick, gray]
table {%
0 0.00861378205128205
400128 0.268429487179487
800000 0.327724358974359
1200128 0.342347756410256
1600000 0.390224358974359
2000128 0.412359775641026
2400000 0.41786858974359
2800128 0.426282051282051
3200000 0.435496794871795
3600128 0.443008814102564
4000000 0.432892628205128
4400128 0.419671474358974
4800000 0.435997596153846
5200128 0.426282051282051
5600000 0.443309294871795
6000128 0.434094551282051
6400000 0.447716346153846
6800128 0.438501602564103
7200000 0.448317307692308
7600128 0.439202724358974
8000000 0.440004006410256
};
\addplot [thick, color1]
table {%
0 0.0107171474358974
400000 0.348157051282051
800000 0.414663461538462
1200000 0.426983173076923
1600000 0.457231570512821
2000002 0.446915064102564
2400002 0.461838942307692
2800031 0.487079326923077
3200031 0.483974358974359
3600034 0.490785256410256
4000005 0.482271634615385
4400087 0.484875801282051
4800036 0.48838141025641
5200017 0.489383012820513
5600006 0.49599358974359
6000130 0.489282852564103
6400103 0.496594551282051
6800115 0.487379807692308
7200157 0.490785256410256
7600112 0.492788461538462
8000043 0.48828125
};
\nextgroupplot[
xlabel={Examples accessed},
ylabel={$m$},
xmin=0, xmax=8000000,
ymin=0, ymax=200,
axis on top,
width=\figurewidth,
height=.5\figureheight,
mystyle
]
\addplot [thick, color1]
table {%
0 16
16000 16
32000 16
48000 16
64000 16
80000 16
96000 16
112000 16
128000 16
144000 16
160000 16
176000 16
192000 16
208000 16
224000 16
240000 16
256000 16
272000 16
288000 16
304000 16
320000 16
336000 16
352000 16
368000 16
384000 16
400000 16
416000 16
432000 16
448000 16
464000 16
480000 16
496000 16
512000 16
528000 16
544000 16
560000 16
576000 16
592000 16
608000 16
624000 16
640000 16
656000 16
672000 16
688000 16
704000 16
720000 16
736000 16
752000 16
768000 16
784000 16
800000 16
816000 16
832000 16
848000 16
864000 16
880000 16
896000 16
912000 16
928000 16
944000 16
960000 16
976000 16
992000 16
1008000 16
1024000 16
1040000 16
1056000 16
1072000 16
1088000 16
1104000 16
1120000 16
1136000 16
1152000 16
1168000 16
1184000 16
1200000 16
1216000 16
1232000 16
1248000 16
1264000 16
1280000 16
1296000 16
1312000 16
1328000 16
1344000 16
1360000 16
1376000 16
1392000 16
1408000 16
1424000 16
1440000 16
1456000 16
1472000 16
1488000 16
1504000 16
1520000 16
1536000 16
1552000 16
1568000 16
1584000 16
1600000 16
1616000 16
1632000 16
1648000 16
1664000 16
1680000 16
1696000 16
1712000 16
1728002 16
1744002 16
1760007 16
1776016 16
1792016 16
1808039 16
1824057 16
1840114 16
1856123 16
1872177 16
1888206 16
1904214 16
1920454 16
1936741 16
1952776 16
1969113 17
1985376 16
2001500 17
2018477 17
2035023 16
2051229 16
2068306 17
2084934 17
2101590 17
2119798 19
2137647 18
2154805 18
2172877 19
2190918 18
2209182 19
2228481 19
2247210 18
2266633 20
2286326 19
2305824 18
2326698 21
2346808 20
2367750 20
2389447 22
2411593 23
2434777 23
2457190 23
2480726 24
2504400 24
2529849 25
2554620 26
2581000 27
2607064 28
2634355 27
2662139 28
2690717 27
2720225 31
2750395 30
2781644 32
2813752 33
2846595 32
2880824 33
2915338 35
2951134 36
2987980 39
3026640 38
3065593 39
3105996 41
3147689 41
3191428 44
3236059 45
3281854 45
3329632 49
3378157 48
3429041 54
3480858 53
3536048 57
3592761 58
3651128 59
3711745 63
3774847 67
3840676 64
3908676 69
3978951 72
4052257 74
4128476 76
4206779 80
4288309 82
4373244 86
4461145 89
4551483 90
4646075 94
4743590 98
4844523 102
4949453 106
5056804 111
5169178 117
5284693 121
5404898 123
5527807 131
5656168 128
5787546 132
5923622 136
6064018 145
6209440 148
6358603 146
6512684 153
6669932 161
6831424 165
6998540 172
7170195 174
7346288 179
7527248 181
7712994 183
7903256 195
};
\end{groupplot}

\end{tikzpicture}
\caption{Additional results for \textsc{cifar-100}.}
\label{fig:results_cifar100_supp}
\end{figure}

\end{document}